\documentclass{article}

   \usepackage[nonatbib, final]{neurips_2022}

\usepackage[numbers]{natbib}
\usepackage[utf8]{inputenc} 
\usepackage[T1]{fontenc}    
\usepackage{hyperref}
\usepackage{url}  
\usepackage{booktabs}
\usepackage{amsfonts}
\usepackage{nicefrac}
\usepackage{microtype}      
\usepackage{xcolor} 
\usepackage{booktabs}
\usepackage{nicefrac}
\usepackage{graphicx}
\usepackage{lipsum}
\usepackage{svg}
\usepackage{adjustbox}
\usepackage{multirow}
\usepackage{amsmath}
\usepackage{amssymb}
\usepackage{pifont}
\usepackage{multicol}
\usepackage{placeins}
\usepackage{xspace}
\usepackage{wrapfig}
\usepackage{floatrow}
\usepackage{enumitem}
\usepackage{caption}
\usepackage{subcaption}
\usepackage{caption}
\newfloatcommand{capbtabbox}{table}[][\FBwidth]
\newcommand\vlpft[1]{\textcolor{gray}{#1}}

\title{
  Few-shot Multimodal Multitask Multilingual Learning \\
  \vspace{0.25em}
}

\author{
  Aman Chadha \\
  Department of Computer Science \\
  Stanford University \\
  \texttt{amanc@stanford.edu} \\
\And
  Vinija Jain \\
  Department of Computer Science \\
  Stanford University \\
  \texttt{vinija@stanford.edu}
}

\begin{document}

\maketitle

\section{Abstract}

While few-shot learning as a transfer learning paradigm has gained significant traction for scenarios with limited data, it has primarily been explored in the context of building unimodal and unilingual models. Furthermore, a significant part of the existing literature in the domain of few-shot multitask learning perform in-context learning which requires manually generated prompts as the input, yielding varying outcomes depending on the level of manual prompt-engineering. In addition, in-context learning suffers from substantial computational, memory, and storage costs which eventually leads to high inference latency because it involves running all of the prompt's examples through the model every time a prediction is made. In contrast, methods based on the transfer learning via the fine-tuning paradigm avoid the aforementioned issues at a one-time cost of fine-tuning weights on a per-task basis. However, such methods lack exposure to few-shot multimodal multitask learning. In this paper, we propose \textbf{f}ew-shot learning for a \textbf{m}ultimodal \textbf{m}ultitask \textbf{m}ultilingual (FM3) setting by adapting pre-trained vision and language models using task-specific hypernetworks and contrastively fine-tuning them to enable few-shot learning. FM3's architecture combines the best of both worlds of in-context and fine-tuning based learning and consists of three major components: (i) multimodal contrastive fine-tuning to enable few-shot learning, (ii) hypernetwork task adaptation to perform multitask learning, and (iii) task-specific output heads to cater to a plethora of diverse tasks. FM3 learns the most prominent tasks in the vision and language domains along with their intersections, namely visual entailment (VE) \cite{xie2019visual}, visual question answering (VQA) \cite{goyal2017making}, and natural language understanding (NLU) tasks such as neural entity recognition (NER) and the GLUE benchmark \cite{wang2018glue} including QNLI \cite{rajpurkar2016squad}, MNLI \cite{williams2017broad}, QQP \cite{quora}, and SST-2 \cite{socher2013recursive}.

\section{Introduction}

Self-supervised pretraining has propelled the adoption of deep learning on tasks with limited labeled data. With their task-agnostic features and improved data efficiency, self-supervised pre-trained models have drastically reduced the opportunity cost to tackle tasks that earlier required a significant amount of data and thus proved intractable using supervised learning. As a result of the advancements in self-supervised pretraining, semi-supervised approaches that combine self-supervision with supervised learning on a task-specific dataset that tackles a related task, have emerged as a new paradigm that has enabled transfer learning. 

One of the biggest open challenges for machine learning research is building models that can be rapidly adapted to novel tasks using only a handful of annotated examples. The domain of few-shot learning (FSL), which is a specific variant of transfer learning, has emerged as an attractive solution to label-scarce scenarios where data annotation can be time-consuming and costly. These methods are designed to work with a small number of labeled training examples, and typically involve adapting pre-trained models for specific downstream tasks. Several flavors of FSL methods exist, each with its pros and cons. 

One such large-scale self-supervised approach, popularized by the arrival of the generative pre-trained transformer (GPT) series \cite{radford2019language, brown2020language} of NLP models, is transfer learning via in-context learning (ICL) which emerges from training at scale. ICL teaches a model to perform a downstream task by feeding in a prompt with a nominal set of supervised examples as input to the model along with a single unlabeled example for which a prediction is desired. In effect, few-shot prompting using a small collection of input-target pairs offers a walk-through to the model on how to transform the input into the output. Notably, since ICL requires no parameter updates, i.e., no gradient-based training is required, a single model can effectively act as a swiss-army knife by being able to immediately perform a wide variety of tasks. ICL, therefore, solely relies on the capabilities that a model learned during pretraining. The ease of use and quick adaptability to target tasks are characteristic features that have caused widespread adoption of ICL \cite{min2021metaicl, lampinen2022can, lazaridou2022internet, min2022rethinking, wang2022benchmarking}. 

While ICL offers a multitude of benefits, it also suffers from several major drawbacks. First, processing all the prompted input-target pairs every time the model makes a prediction incurs significant compute, memory, and latency costs. These costs stack up as the number of the inferences increases -- in a situation where the goal is to perform inference over a batch of test examples rather than one-off predictions, ICL can prove to be impractical from a resource standpoint. Second, owing to a limited-length context window, the number of support examples $k$ that the model can utilize are restricted to nominal numbers. This is because we must fit all $k$ examples into the model's context window, which is limited to a specific number of tokens (1024 in case of GPT-2 and 2048 in case of GPT-3). Third, ICL typically produces inferior performance compared to fine-tuning \cite{alayrac2022flamingo, radford2019language, liu2022few}. Finally, while the model's performance is a function of semantic and structural aspects of the prompt which can cause a significant yet unpredictable impact on the model’s performance \cite{lu2021fantastically}, far beyond inter-run variation of fine-tuning. In particular, semantic changes such as the phrasing or choice of words in the prompt and syntactic changes such as the exact formatting of the prompt (including the wording \cite{webson2021prompt} and ordering of examples \cite{zhao2021calibrate}) can cause a significant, unintended, and difficult-to-estimate impact on the model's performance. Furthermore, recent work has also demonstrated that ICL can perform well even when provided with incorrect labels, raising concerns as to how much learning is taking place at all \cite{min2022rethinking}. 

Another common semi-supervised learning paradigm is transfer learning via fine-tuning (FT) which follows a two-staged process: (i) utilize the parameters of a pre-trained large-scale self-supervised model learning for weight initialization, and (ii) perform gradient-based fine-tuning using data associated with the downstream task of interest. With the advent of representation-learning approaches such as BERT \cite{devlin2018bert}, the domain of NLP underwent a radical transformation from supervised to semi-supervised approaches for tasks such as sentiment analysis, neural entity recognition, question answering, summarization, conversational response generation, etc. Representation-learning approaches have now taken center-seat in NLP, with the learned contextualized representations from these pre-trained models serving as initial task-agnostic features that, in turn, offer a the starting point for learning task-specific features. While problems with limited labeled data have benefited significantly owing to the reduced data-appetite of semi-supervised approaches, tasks with abundant labeled data have also seen improved performance.

While FT has produced many state-of-the-art (SoTA) results \cite{sanh2021multitask} on a range of classification tasks, it results in a model that is specialized for a single task with an entirely new set of parameter values, which can become impractical when FT a model on many downstream tasks. In other words, such models typically perform one task at a time, and cannot learn new concepts or adapt to new tasks in a few shots. FM3 seeks to address this gap and enable multimodal FSL -- much like how \textsc{SetFit} contrastively fine-tunes pre-trained Sentence Transformer models \cite{reimers-2019-sentence-bert} and dispenses with prompts altogether and does not require large-scale pre-trained LMs to achieve high accuracy. With only 8 labeled examples in the Customer Reviews (CR) sentiment dataset, \textsc{SetFit} is competitive with RoBERTa finetuned on the full training set \cite{tunstall2022efficient}, despite the fine-tuned model being three times larger. 

Both ICL and fine-tuning have been explored in a multimodal context. A slew of methods, notably \textit{Flamingo} \cite{alayrac2022flamingo} and \textit{Frozen} \cite{tsimpoukelli2021multimodal}, perform ICL with the final objective to have the model rapidly adapt to a variety of multimodal tasks. While \textit{Flamingo} achieves competitive performance with FSL, in some cases outperforming models fine-tuned on thousands of times more task-specific data, \textit{Frozen} offers relatively lower performance in return for the flexibility of using an off-the-shelf pre-trained LM and keeping its weights frozen. On the other hand, Oscar \cite{li2020oscar} and Omninet \cite{pramanik2019omninet} are multimodal multitask models that do not perform ICL. While Oscar is pre-trained using pre-trained with aligned data on task-agnostic cross-modal objectives (a masked token loss over words and visual tags, and a contrastive loss between visual tags and others) and then fine-tuned to specific tasks, Omninet is simultaneously trained on its target tasks and undergoes no finetuning. In the zero-/few-shot learning context, multimodal pretraining has recently shown to enable strong generalization in the discriminative setting using large-scale contrastive learning \cite{radford2021learning, jia2021scaling}.

An additional paradigm for enabling a model to perform a new task with minimal updates is parameter efficient fine-tuning (PEFT), where a pre-trained model is fine-tuned by only updating a small number of added or selected parameters. Recent methods have matched the performance of fine-tuning the full model while only updating or adding a small fraction (e.g. 0.01\%) of the full model’s parameters
\cite{hu2021lora, lester2021power}. Furthermore, certain PEFT methods allow mixed-task batches where different examples in a batch are processed differently \cite{lester2021power}, making both PEFT and ICL viable for multitask models. While the benefits of PEFT address some shortcomings of fine-tuning (when compared to ICL), there has been relatively little focus on whether PEFT methods work well when very little labeled data is available. \cite{liu2022few} closes this gap by proposing T-Few, a model that learns using PEFT and a fixed set of hyperparameters, attaining strong performance on novel, unseen tasks while only updating a tiny fraction of the model’s parameters. 

FM3 combines the best of both worlds of ICL- and FT-based transfer learning and offers an efficient and prompt-free framework that offers strong generalization to new multimodal vision-language tasks in a few-shot setting. Despite the flexibility offered by ICL, its limitations leave much to be desired, especially in situations where compute, latency, memory, batch inference, performance determinism, etc. are important. On the other hand, FT offers performance invariance since it does not require prompts, offers better performance than ICL-based methods \cite{alayrac2022flamingo}, and is resource-efficient in terms of compute, latency, memory, etc. While zero-/few-shot generalization is a desirable by-product of ICL, the only significant downside to FT is that generalization to new tasks with limited data is challenging. FM3 is architected keeping the aforementioned drawbacks of ICL in mind and thus follows the FT approach but overcomes its limitations as follows: (i) multimodal contrastive fine-tuning to enable FSL, (ii) using hypernetworks with a limited parameter count to perform task adaptation for multitask learning, and (iii) task-specific output heads to cater to a plethora of diverse tasks.

FM3 achieves high accuracy with little labeled data - for instance, with only 16 labeled examples per class on the complex task of SNLI-VE \cite{xie2019visual}, FM3 surpasses the current SoTA fine-tuned on the full training set of 430K examples! 
Compared to other FSL methods, FM3 has several unique features:
\begin{itemize}
    \item \textbf{No prompts or verbalisers:} Current techniques for few-shot fine-tuning require handcrafted prompts or verbalisers to convert examples into a format that's suitable for the underlying language model. FM3 dispenses with prompts altogether by generating rich embeddings directly from text examples. This obliterates the need for manual prompt engineering, which in turn, results in performance determinism.
    \item \textbf{Resource efficiency:} Optimal use of compute, latency, memory, etc. compared to our baselines \textit{Flamingo}, \textit{Frozen}, and especially ICL-based methods.
    \item \textbf{Frozen pre-trained models:} FM3 uses pre-trained vision and language encoders without fine-tuning them. This implies that FM3 architecture enables drop-in plug-and-play replacement for modality encoders. Only small hypernetwork models need to be fine-tuned when experimenting with different encoders.
    \item \textbf{Fast to train:} FM3 doesn't require large models like \textit{Flamingo} (80B) or \textit{Frozen} (7B+) to achieve high accuracy. As a result, it is significantly faster to train and run inference with.
    \item \textbf{Multilingual support:} FM3 enables multilingual processing, and can be paired up with any multilingual text encoder such as multilingual Sentence Transformer \cite{reimers-2019-sentence-bert} variants of MPNet \cite{song2020mpnet}, RoBERTa \cite{liu2019roberta}, ALBERT \cite{lan2019albert}, LASER \cite{feng2020language}, etc., which enables multilingual learning in 50+ languages by simply fine-tuning a multilingual model checkpoint.
\end{itemize}

While proposals that address a subset of the areas of few-shot multimodal multitask multilingual learning exist, to our knowledge, FM3 is the first to explore the intersection of the domains of multimodal multitask multilingual learning in a FSL setting.

\section{Related Work}

\subsection{Few-shot learning using pre-trained models}

In the domain of NLP, \textsc{SetFit} proposed by Tunstall et al. \cite{tunstall2022efficient} is an efficient and prompt-free framework for few-shot fine-tuning of Sentence Transformers (ST). \textsc{SetFit} works by fine-tuning a pre-trained ST on a small number of text pairs in a contrastive Siamese manner. The resulting model is then used to generate rich text embeddings, which are used to train a classification head. This simple framework requires no prompts, and achieves high accuracy with orders of magnitude less parameters than existing techniques. \textsc{SetFit} obtains comparable results with parameter-efficient fine-tuning (PEFT) and parameter efficient tuning (PET) techniques, while being an order of magnitude faster to train. \textsc{SetFit} achieves high accuracy with little labeled data - for instance, with only 8 labeled examples per class on the Customer Reviews sentiment dataset \cite{mcauley2013hidden}, \textsc{SetFit} is competitive with fine-tuning RoBERTa Large on the full training set of 3K examples. Owing to its practical utility in enabling FSL, we adopt the idea of contrastive fine-tuning from \textsc{SetFit} and generalize it to a multimodal multitask multilingual setting as part of FM3.

\subsection{Multitask fine-tuning using PEFT}

In \cite{houlsby2019parameter}, Houlsby et al. propose a parameter-efficient fine-tuning method which introduces adapter modules between the layers of a pre-trained language model. Adapter modules yield a compact and extensible model; they add only a few trainable parameters per task, and new tasks can be added without revisiting previous ones. The parameters of the original network remain fixed, yielding a high degree of parameter sharing. They achieve SoTA performance on GLUE \cite{wang2018glue} whilst adding only a few parameters per task. However, the downside of this approach is that they are trained separately for each task and thus do not enable sharing information across tasks. 

To circumvent the aforementioned issue of knowledge sharing across tasks, Mahabadi et al. \cite{mahabadi2021parameter} learn adapter parameters for all layers and tasks using shared hypernetworks, which condition on task, adapter position, and layer ID in a transformer model. This parameter-efficient multitask learning framework achieves the best of both worlds by sharing knowledge across tasks via hypernetworks while enabling the model to adapt to each individual task through task-specific adapters. Experiments on the GLUE benchmark show improved performance in multitask learning while adding only 0.29\% parameters per task. Given the fact that hypernetworks enable easy multitask fine-tuning of pre-trained models without having to actually fine-tune the model's weights (i.e., they remain frozen in this process), we adopt multitask finetuning in our proposed architecture.

\subsection{Multitask multimodal learning}

Hu and Singh propose UniT \cite{hu2021unit}, a Unified Transformer encoder-decoder model that learns 7 tasks jointly across 8 datasets spread over different vision and lanuage domains, ranging from object detection to natural language understanding and multimodal reasoning. UniT achieves strong performance with significantly few parameters in some cases outperforming separately trained single task models. While the architecture offers joint end-to-end training of each task, it requires a substantial amount of data across all tasks for the model to generalize. Our approach, on the other hand, utilizes FSL to efficiently learn a task with a small fraction of data. 

In \cite{pramanik2019omninet}, Pramanik et al. propose OmniNet, a single model to support tasks with multiple input modalities as well as asynchronous multitask learning. OmniNet is powered by a spatio-temporal cache that enables learning spatial dimension of the input in addition to the hidden states corresponding to the temporal input sequence. Even though OmniNet is $3\times$ parameter-efficient, there is a significant performance gap on most tasks it was trained on compared to the individual model counterparts.

\subsection{Multitask multilingual multimodal learning}

$M^{3}P$, proposed in \cite{ni2021m3p}, is a multitask multilingual multimodal pre-trained model that combines multilingual pre-training and multimodal pre-training into a unified framework via multitask pre-training. $M^{3}P$ learns universal representations that can map objects occurred in different modalities or texts expressed in different languages into a common semantic space. In addition, to alleviate the issue of lack of sufficient labeled data for non-English multimodal tasks, they propose multimodal code-switched training (MCT) \cite{qin2020cosda} which replaces each word in the caption with a translated word with a probability of $\beta$. If a word has multiple translations, a random one is chosen. Experiments on the multilingual image retrieval task across MS COCO \cite{lin2014microsoft} and Multi30K \cite{elliott2016multi30k} show competitive results for English and new establish SoTA results for non-English languages. While $M^{3}P$ tackles a similar problem as FM3
, it does not assume any restrictions on the annotation budget -- in other words, it does not consider the few-shot setting for learning its tasks.

\subsection{Few-shot multimodal multitask learning}

In \cite{alayrac2022flamingo}, Alayrac et al. introduce Flamingo, a family of Visual Language Models (VLM) trained on large-scale multimodal web corpora with an ability to rapidly adapt to a variety of image and video tasks. \textit{Flamingo} proposes the following key architectural innovations: (i) bridge powerful pre-trained vision-only and language-only models, (ii) handle sequences of arbitrarily interleaved visual and textual data, and (iii) seamlessly ingest images or videos as inputs. The end result is a single \textit{Flamingo} model that can achieve a new SoTA with FSL, simply by prompting the model with task-specific examples. On numerous benchmarks, \textit{Flamingo} outperforms models fine-tuned on thousands of times more task-specific data. These include open-ended tasks such as visual question-answering, where the model is prompted with a question which it has to answer; captioning tasks, which evaluate the ability to describe a scene or an event; and close-ended tasks such as multiple-choice visual question-answering.

In \cite{tsimpoukelli2021multimodal}, Tsimpoukelli et al. present \emph{Frozen}, a simple-yet-effective approach for transferring the FSL abilities inherent in large auto-regressive language models to a multimodal setting (vision and language). \emph{Frozen} is a multimodal few-shot learner, with the surprising ability to learn a variety of new tasks when conditioned on examples, represented as a sequence of multiple interleaved image and text embeddings. \emph{Frozen} can rapidly learn words for new objects and novel visual categories and do visual question-answering with only a handful of examples. While this work serves as an important baseline for FM3, a key limitation is that it achieves far from SoTA performance on the specific tasks that it learns in a few shot setting. \emph{Frozen} shows that training a visual encoder through a pre-trained and frozen language model results in a system capable of strong out-of-distribution (zero-shot) generalization. Furthermore, \emph{Frozen} confirms that the ability to rapidly adapt to new tasks given appropriate prompts is inherited from the pre-trained language model and transfers directly to multimodal tasks.

While \textit{Flamingo} and \emph{Frozen} are both ICL-based FSL methods, the differentiating factors are: (i) the scale of data used to train these models, and (ii) architectural variations. \textit{Flamingo} is trained on large-scale multimodal web corpora while is \emph{Frozen} is trained on the Conceptual Captions dataset \cite{sharma2018conceptual}. The architectural design choices differ between the two in using pre-trained modality encoders vs. training them from scratch. Similar to FM3, \textit{Flamingo} uses off-the-shelf pre-trained encoders and only generates adapter components (in the form of Perceiver Resampler blocks) while \emph{Frozen} utilizes a pre-trained LM but trains its own vision encoder that feeds the LM. Inspired by this observation, FM3 borrows the idea of using separate text and vision adapters in the form of hypernetworks so as to offer the model additional degrees of freedom, which in turn, helps render better performance. 

\section{FM3}

\begin{figure}[!h]
    \centering
    \includegraphics[width=1\linewidth]{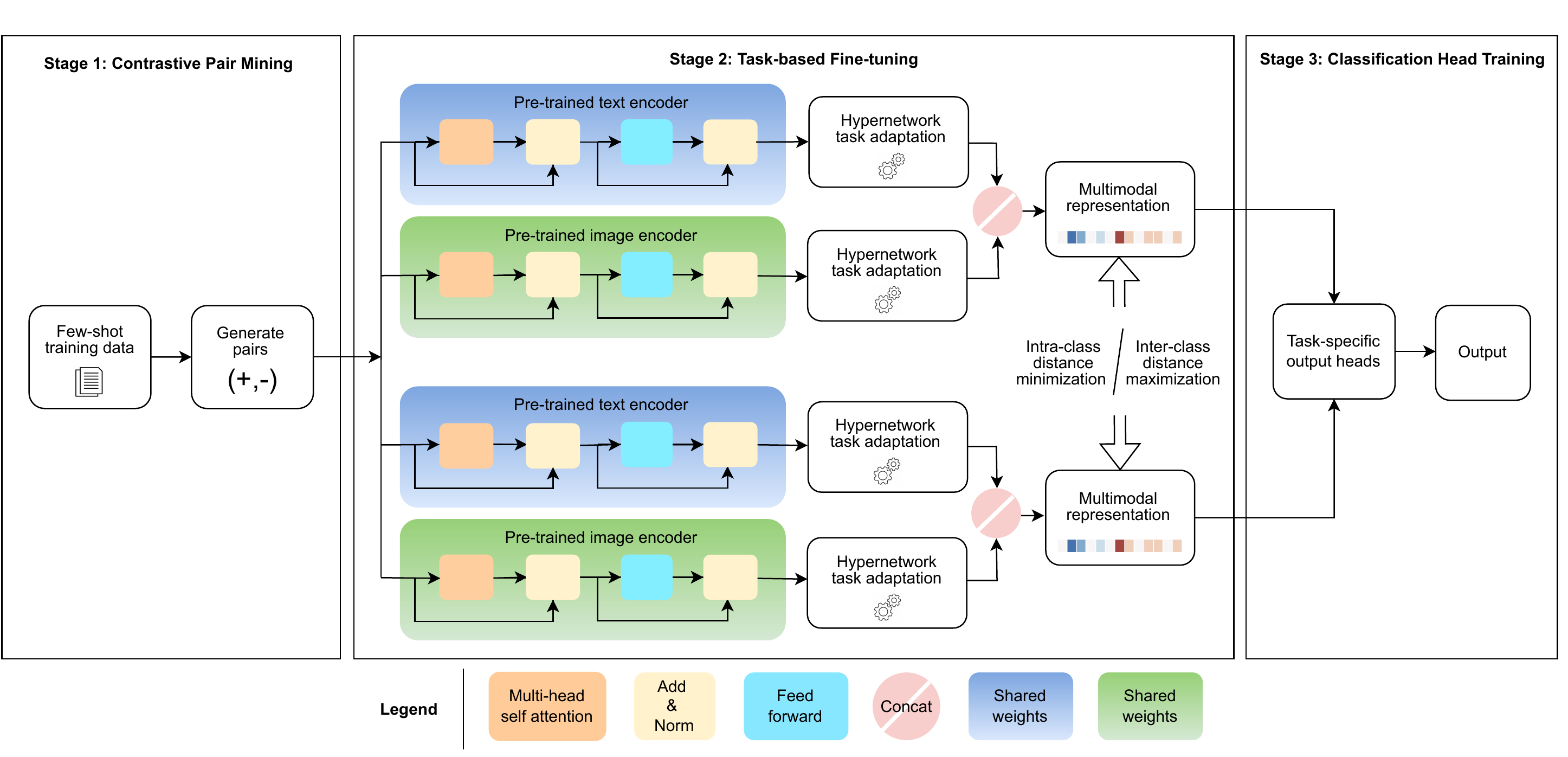} 
    \vspace {-2mm}
    \caption{Architectural overview of FM3.
    FM3 consists of three stages: (i) contrastive pair mining for fine-tuning, which generates positive and negative pairs, (ii) task-based fine-tuning involves adapting the pre-trained text and image encoder models for down-stream tasks using hypernetworks, and (iii) training task-specific classification heads.}
    \label{flow}    
\end{figure}

\subsection{Methods}

Figure \ref{flow} offers a visual summary of the architectural stages of FM3.

\subsubsection{Task and batch sampling}
At each iteration during training, we randomly select a task with a sampling probability that can be manually specified based on the dataset size. Once the task list has been sampled, for tasks with multiple datasets, we randomly sample a dataset corresponding to that task to fill a batch of samples.

\subsubsection{Contrastive fine-tuning for few-shot learning}\label{contrastive}
Similar to \cite{tunstall2022efficient}, we utilize a contrastive learning approach to FSL. Contrastive learning effectively enlarges the size of training data which is critical in few-shot scenarios and thus fosters effective learning for tasks with limited annotated data. Assuming a small number ($k$) of labeled examples for a classification task, the potential size of the fine-tuning set $T$ derived from the number of unique positive and negative contrastive pairs that can be generated would be $\frac{k(k - 1)}{2}$, which is significantly larger than just $k$ \cite{tunstall2022efficient}. In this stage, we sample $R$ positive and $R$ negative triplet pairs, where $R$ is a hyperparameter (set to 20, following \cite{tunstall2022efficient}). 
We utilize multiple negatives ranking loss \cite{henderson2017efficient} for contrastive fine-tuning owing to its superior performance \cite{henderson2017efficient} and its ability to randomly sample negative pairs from each batch in an automated fashion.


\subsubsection{Task-based fine-tuning using hypernetworks}\label{hypernetworks}

To our knowledge, FM3 is the first to utilize hypernetworks in a multimodal setting. Using frozen modality encoders has the distinct advantage of preventing catastrophic forgetting (compared to fine-tuning the encoders themselves) \cite{alayrac2022flamingo}. As such, we utilize an independent hypernetwork for each modality.

In this step, we perform task-specific fine-tuning of SoTA pre-trained text and vision models, namely a pre-trained multilingual MPNet \cite{song2020mpnet} Sentence Transformer \cite{reimers-2019-sentence-bert} from Huggingface \cite{mpnethf} as our text backbone and CoCa-Base \cite{yu2022coca} as our vision backbone. We adopt the idea of hypernetworks from \cite{mahabadi2021parameter}, which is a parameter-efficient method for multitask fine-tuning. We train shared hypernetworks to generate task-specific adapters conditioned on the task, layer ID, and adapter position embeddings. These shared hypernetworks capture knowledge across tasks, enabling positive transfer to low-resource and related tasks, while task-specific layers allow the model to adapt to each individual task. 
We optimize a distance function based on cosine similarity, minimizing it for positive pairs and maximizing it for negative pairs. 

\subsubsection{Task-specific classification head training}\label{classification}
Lastly, we train task-specific classification heads on the fine-tuned model obtained from the above step. The generated embeddings corresponding to the data samples for each task, along with their labels, constitute the training set for the respective classification head. We use logistic regression for binary classification tasks such as SST-2, QQP, QNLI, etc. and softmax for multiclass classification tasks such as VQA, SNLI-VE, GLUE, NER, etc.

\subsection{Tasks and Datasets} 

Table \ref{tab:1} delineates the domains, tasks, and datasets for training and evaluating FM3.

\begin{table*}[ht]
    \caption{Datasets for training and evaluation}
    \label{tab:1}
\begin{adjustbox}{width=1.00\textwidth,center}    
    \begin{tabular}{ccc}
    \toprule
    Domain & Task & Dataset \\
    \hline
\multirow{2}{*}{Language understanding} & Neural entity recognition (NER) & CoNLL-2003 \cite{sang2003introduction} \\
& GLUE benchmark \cite{wang2018glue} & QNLI \cite{rajpurkar2016squad}, MNLI \cite{williams2017broad}, QQP \cite{quora}, and SST-2 \cite{socher2013recursive} \\
\hline
\multirow{2}{*}{Vision-and-language reasoning} & Visual entailment & SNLI-VE \cite{xie2019visual} \\ 
& Visual question answering (VQA) & VQAv2 dataset \cite{goyal2017making} (with questions from Visual Genome \cite{krishna2017visual} as additional data), OK-VQA \cite{marino2019ok} \\
 \bottomrule
    \end{tabular}
\end{adjustbox}    
\end{table*}

\section{Experiments}\label{expts}

\subsection{Experimental setup}

We finetuned FM3 on the Conceptual Captions dataset (which \textit{Frozen} \cite{tsimpoukelli2021multimodal} is trained on) for vis-à-vis comparisons. We use the AdamW optimizer with global norm clipping of 1, no weight decay for the hypernetworks and weight decay of 0.1 for the other trainable parameters. We anneal the learning rate, increasing it linearly from 0 to $10^{-3}$ up over the first 5000 steps then held constant for the duration of training and then decayed exponentially. Unless specified otherwise we train our models for 500K steps. All datasets were trained with the same weights. Since the performance of models trained with a contrastive objective is sensitive to the batch size, we use a relatively large batch size of 32.

\subsection{Baselines}

We utilize \textit{Flamingo} \cite{alayrac2022flamingo} and \emph{Frozen} \cite{tsimpoukelli2021multimodal} as our primary baselines since they deal with multimodal multitask learning in the context of FSL. We include UniT \cite{hu2021unit} as an additional baseline since it deals with multimodal multitask learning of tasks that \textit{Flamingo} \cite{alayrac2022flamingo} and \emph{Frozen} haven't been evaluated on. For each task, we compare FM3 with both task-specific zero-/few-shot and pre-trained/fine-tuned SoTA. Since none of the above baselines support multilingual tasks, we utilize \cite{ni2021m3p} as a baseline to qualify the performance of FM3 on non-English tasks.

\subsection{Results}

\begin{table}[!h]
\resizebox{\textwidth}{!}{%
\begin{tabular}{ccccccccccccccccccc}
Method & FT & Shot(s)     & 
\rotatebox[origin=c]{90}{OKVQA \cite{marino2019ok}}     & \rotatebox[origin=c]{90}{VQAv2 \cite{goyal2017making}}     & 
\rotatebox[origin=c]{90}{Flickr30K \cite{plummer2015flickr30k}} & 
\rotatebox[origin=c]{90}{Multi30K \cite{elliott2016multi30k}} & 
\rotatebox[origin=c]{90}{SNLI-VE \cite{xie2019visual}} & 
\rotatebox[origin=c]{90}{CoNLL-2003 \cite{sang2003introduction}} & 
\rotatebox[origin=c]{90}{QNLI \cite{rajpurkar2016squad}} & 
\rotatebox[origin=c]{90}{MNLI \cite{williams2017broad}} & 
\rotatebox[origin=c]{90}{QQP \cite{quora}} &
\rotatebox[origin=c]{90}{SST-2 \cite{socher2013recursive}} &
\\ \hline

\begin{tabular}[c]{@{}c@{}}Zero/Few \\ shot SoTA\end{tabular}& \ding{55} & Various & \begin{tabular}[c]{@{}c@{}}\small{[\citenum{gui2021kat}]}\\ 43.3 \\ (16)\end{tabular}     & \begin{tabular}[c]{@{}c@{}}\small{[\citenum{tsimpoukelli2021multimodal}]}\\ 38.2\\ (4)\end{tabular} & \begin{tabular}[c]{@{}c@{}}\small{[\citenum{wang2022image}]}\\94.9 \\(0)\end{tabular} & - & 
\begin{tabular}[c]{@{}c@{}}\small{[\citenum{vu2021strata}]}\\87.3 \\(0)\end{tabular} & 
\begin{tabular}[c]{@{}c@{}}\small{[\citenum{huang2021few}]}\\65.4 \\(5)\end{tabular} & 
- & 
\begin{tabular}[c]{@{}c@{}}\small{[\citenum{vu2021strata}]}\\86.4 \\(5)\end{tabular} & 
\begin{tabular}[c]{@{}c@{}}\small{[\citenum{zhang2021differentiable}]}\\67.8 \\(16)\end{tabular} & 
\begin{tabular}[c]{@{}c@{}}\small{[\citenum{zhang2021differentiable}]}\\93.5 \\(16)\end{tabular} & 
\\ \hline

\multirow{3}{*}{Flamingo-80B} & \multirow{3}{*}{\ding{55}} & 0 & 50.6    & 56.3 & 67.2 & \multirow{3}{*}{-} & \multirow{3}{*}{-} & \multirow{3}{*}{-} & \multirow{3}{*}{-} & \multirow{3}{*}{-} & \multirow{3}{*}{-} & \multirow{3}{*}{-}\\
& & 4 & 57.4    & 63.1 & 75.1 & & & & & & & & \\
& & 32& {57.8} & {67.6}  & {75.4} & & & & & & & & \\
\hline  

\multirow{3}{*}{\textit{Frozen}} & \multirow{3}{*}{\ding{55}} & 0 & 5.9    & 29.5 & \multirow{3}{*}{-} & \multirow{3}{*}{-} & \multirow{3}{*}{-} & \multirow{3}{*}{-} & \multirow{3}{*}{-} & \multirow{3}{*}{-} & \multirow{3}{*}{-} & \multirow{3}{*}{-}\\
& & 1 & 9.7 & 35.7 \\
& & 4 & {12.6} & {38.2} \\
\hline

{UniT} & \ding{55} & \begin{tabular}[c]{@{}c@{}}Fully\\ Supervised\end{tabular} & -    & 67.0 & - & - & 73.2 & - & 88.0 & 81.8 & 90.6 & 91.5 \\
\hline

\multirow{4}{*}{\textbf{FM3-5B}} & \multirow{4}{*}{\ding{55}} & 0 & 51.5 & 57.3 & 93.2 & 27.2 & 86.2 & 51.3 & 53.6 & 58.2 & 41.4 & 75.5 \\
& & 4 & 55.5 & 64.6 & 94.3 & 35.2 & 89.9 & 66.2 & 67.6 & 55.5 & 54.4 & 86.8 \\
& & 16 & 57.8 & {67.9}  & {96.1} & 37.1 & 92.4 & 77.3 & 75.4 & 81.1 & 66.7 & 95.7 \\
& & 64 & \underline{\textbf{58.9}} & \underline{\textbf{71.2}}  & \underline{\textbf{99.1}} & 39.9 & \underline{\textbf{94.4}} & \textbf{82.7} & 83.7 & 86.4 & \textbf{78.9} & \underline{\textbf{99.2}} \\
\hline

\begin{tabular}[c]{@{}c@{}}\vlpft{Pre-trained}\\  \vlpft{FT SoTA}\end{tabular} & \vlpft{\ding{52}} & Various & \begin{tabular}[c]{@{}c@{}}\vlpft{54.4}\\ \vlpft{\small{[\citenum{gui2021kat}]}}\\ \vlpft{(10K)}\end{tabular} & \begin{tabular}[c]{@{}c@{}}\vlpft{80.2}\\ \vlpft{\small{[\citenum{yuan2021florence}]}}\\ \vlpft{(444K)}\end{tabular} & 
\begin{tabular}[c]{@{}c@{}}\vlpft{98.8}\\ \vlpft{\small{[\citenum{zeng2022x}]}}\\ \vlpft{(31K)}\end{tabular} & 
\begin{tabular}[c]{@{}c@{}}\vlpft{49.3}\\ \vlpft{\small{[\citenum{shan2022ernie}]}}\\ \vlpft{(31K)}\end{tabular} & 
\begin{tabular}[c]{@{}c@{}}\vlpft{91.0}\\ \vlpft{\small{[\citenum{wang2022unifying}]}}\\ \vlpft{(430K)}\end{tabular} & 
\begin{tabular}[c]{@{}c@{}}\vlpft{94.6}\\ \vlpft{\small{[\citenum{wang2020automated}]}}\\ \vlpft{(430K)}\end{tabular} &  
\begin{tabular}[c]{@{}c@{}}\vlpft{99.2}\\ \vlpft{\small{[\citenum{lan2019albert}]}}\\ \vlpft{(105K)}\end{tabular} &  
\begin{tabular}[c]{@{}c@{}}\vlpft{92.0}\\ \vlpft{\small{[\citenum{raffel2020exploring}]}}\\ \vlpft{(393K)}\end{tabular} &  
\begin{tabular}[c]{@{}c@{}}\vlpft{89.2}\\ \vlpft{\small{[\citenum{wang2021entailment}]}}\\ \vlpft{(364K)}\end{tabular} &  
\begin{tabular}[c]{@{}c@{}}\vlpft{97.5}\\ \vlpft{\small{[\citenum{raffel2020exploring}]}}\\ \vlpft{(67K)}\end{tabular}
 \\ 
\bottomrule
\end{tabular}
}

\vspace{0.5em}

\caption{
\textbf{Comparison to the state of the art.} A \emph{single} FM3 model reaches the state of the art on a wide array of vision-language understanding tasks with FSL, significantly outperforming previous best zero- and few-shot methods with as few as 16 examples.
More importantly, using only $64$ examples and without adapting any model weights, FM3 {\em outperforms} the current best methods -- fine-tuned on thousands of annotated examples -- on 4 tasks.
Best few-shot numbers across all shots are in \textbf{bold}, best numbers across both zero/few-shot (prompt based) and fine-tuned models are {\underline{underlined}}. For each baseline, we chose the best numbers across spanning all variants/experiments (unless explicitly stated).
\label{tab:res}
}
\vspace{-0.5cm}
\end{table}

Table \ref{tab:res} performs a comparative analysis of FM3 with \textit{Flamingo}, \textit{Frozen}, and the respective zero-/few-shot and fine-tuned SoTA on each task with number of support examples/shots as $k \in \{0, 4, 16, 64\}$. While Flickr30K Image-to-Text uses Recall@1, Mulit30K \texttt{en-de} uses BLEU, CoNLL-2003 and QQP use F1, all other tasks utilize accuracy as their performance metric. Note that since \textit{Frozen} is an auto-regressive model/decoder which undergoes prompt-based fine-tuning, $k$ indicates the number of support examples as part of the prompt/prefix passed as input to the model, while FM3 being an encoder-based architecture, $k$ indicates the number of examples we contrastively fine-tune on.



\textbf{Few-shot results.} FM3 outperforms zero-/few-shot baselines on 7 out of the 10 benchmarks considered. This is achieved with as few as 16 examples per task, demonstrating superior adaptation to these tasks. More importantly, FM3 is often competitive with SoTA methods fine-tuned on up to hundreds of thousands of annotated examples. On 4 out of 10 tasks, FM3 even outperforms the fine-tuned SoTA despite using a single set of model weights and only 64 task-specific examples.

\textbf{Scaling with respect to parameters and shots.} As table \ref{tab:res} indicates, more the number of shots, better the few-shot performance, similar to GPT-3 \cite{radford2019language}. The performance improvement shows diminishing returns as the number of shots increases.

\subsection{Inference runtime analysis}

Table \ref{tab:inftime} summarizes our inference runtime analysis. We measure the time taken to run FM3 and our primary baselines on the test set of VQAv2 \cite{goyal2017making} and OKVQA \cite{marino2019ok} and averaging it out by the total number of total samples. These measurements are from a platform with NVIDIA A100 with 32GB VRAM. \textbf{Bold} numbers indicate best performance. \underline{Underlined} numbers indicate the next best baseline on which the \% improvements for FM3 are based.

\begin{table}[!h]
\hspace*{-0.25cm}
\resizebox{0.5\columnwidth}{!}{%
\begin{tabular}{lcc}
& \textbf{VQAv2} (sec.) & \textbf{OKVQA} (sec.) \\ \toprule
\textbf{\textbf{FM3}} & \textbf{0.187} (58\% $\uparrow$) & \textbf{0.214} (46\% $\uparrow$) \\
\midrule
\text{\textit{Flamingo}} & 0.353 & 0.371 \\
 \midrule
\text{\textit{Frozen}} & \underline{0.295} & \underline{0.312} \\
\bottomrule
\end{tabular}
}
\captionsetup{width=.5\textwidth}
\caption{Inference runtime analysis of FM3 vs. \textit{Flamingo} and \textit{Frozen} on VQAv2 and OKVQA. Measurements are based on wall clock time (sec.) so lower is better.}
\label{tab:inftime}
\end{table}

\section{Ablation analysis}

Table \ref{abl} summarizes the results for FM's ablation experiments. \textbf{Bold} numbers indicate best performance. \underline{Underlined} numbers indicate the baseline on which the \% numbers are based. We analyze the impact of the following design decisions on FM3's performance: 

\begin{itemize}
    \item \textbf{Direct encoder fine-tuning with no hypernetworks.} While frozen modality encoders prevents catastophic forgeting \cite{alayrac2022flamingo}, we adopt a hypernetwork-fee architecture and fine-tune the encoders themselves with data from our target tasks. 
    \item \textbf{Hypernetwork size selection.} We perform comparisons with varied parameter size allocations for hypernetworks to quantify the effect of hypernetwork size. The parameter count for hypernetworks is expressed as a percentage of the baseline FM3 model parameter count. 
    \item \textbf{Compute/memory vs. performance trade-offs.} We vary the choice of text and vision encoders (which in turn, varies the number of parameters and time complexity of the model). For our text encoder, we choose mulitingual MiniLM \cite{wang2020minilm} with 57\% lesser parameters compared to our default choice of the \texttt{paraphrase-multilingual-mpnet-base-v2} variant of multilingual MPNet \cite{mpnethf}. For our vision encoder, we choose the \texttt{vit-base-patch16-224} variant of ViT \cite{dosovitskiy2020image} with 78\% fewer parameters compared to our default choice of CoCa-Base \cite{yu2022coca}.

\end{itemize}

\begin{table}[!h]
\hspace*{-0.25cm}
\capbtabbox{
\begin{tabular}{lcc}
& \textbf{VQAv2} (acc.) & \textbf{OKVQA} (acc.) \\ \toprule
\textbf{\textbf{FM3 (default)}} & \underline{\textbf{71.2}} & \underline{\textbf{58.9}} \\
\midrule
\text{{No hypernetworks}} & 64.3 (90.0\% $\downarrow$) & 52.1 (88.4\% $\downarrow$) \\
\text{{Hypernetworks with 5\% parameters}} & {69.5} (97.6\% $\downarrow$) & {56.7} (88.4\% $\downarrow$) \\
\text{{Text encoder: MiniLM}} & {68.1} (95.6\% $\downarrow$) & {55.4} (94.0\% $\downarrow$) \\
\text{{Vision encoder: ViT}} & {66.2} (92.9\% $\downarrow$) & {51.2} (86.9\% $\downarrow$) \\
\text{{Text/vision encoders: MiniLM/ViT}} & {65.5} (91.9\% $\downarrow$) & {52.3} (88.7\% $\downarrow$) \\
\bottomrule
\end{tabular}
}{
\caption{Ablation analysis of FM3 on VQAv2 and OKVQA with number of shots as 64. Default FM3 uses hypernetworks with 10\% parameters, and MPNet/CoCa as text/vision encoders. Measurement units are \% accuracy so higher is better.}
\label{abl}
}
\end{table}

\section{Future Work}

While FM3 establishes a new SoTA on several tasks, there are significant opportunities for improvement centered around three major aspects: (i) data, (ii) model architecture, and (iii) loss function. First, \textit{Flamingo} \cite{alayrac2022flamingo} highlights the importance of a diverse dataset amalgamated from various disparate sources (\textit{Flamingo} uses $>$2B image-text pairs vs. 3.3M that FM3 was trained on) in training the neural network. Using the publicly available massive LAION-400M dataset \cite{schuhmann2021laion} would be a great starting point. Second, the model architecture can incorporate other techniques that offer reasonably high performance with a reduced parameter count such as low-rank based adaption methods, for e.g., LoRA \cite{hu2021lora}.
Third, following \cite{qi2020imagebert, li2020oscar}, we can formulate the ranking loss \cite{henderson2017efficient} as a binary classification problem. This has reported to lead to an increase in performance \cite{qi2020imagebert, li2020oscar}. In other words, given an aligned image-text pair, we randomly select a different image or a different caption to form an unaligned pair. Similar to FM3's current framework, the final concatenated multimodal embedding can still be used as the input for classification to predict whether the given pair is aligned or not. Finally, FM3 is easily extendable to other languages, tasks, and modalities. 

\section{Conclusion}

FM3 combines the best of both worlds of in-context learning and fine-tuning as a front-runner in the niche domain of few-short multilingual multimodal multitask learning. It offers a scalable architecture that can span modalities, tasks, and languages, all while setting a new standard with SoTA performance on a plethora of tasks and competitive performance on others. FM3 outperforms zero-/few-shot baselines on 7 out of 10 benchmarks with as few as 16 examples per task. Moreover, FM3 is competitive with fine-tuning a plethora of task-specific SoTA models on fine-tuned on up to hundreds of thousands of annotated examples. On 4 out of 10 tasks, FM3 even outperforms the fine-tuned SoTA despite using a single set of model weights and only 64 task-specific examples. Lastly, FM3 also yields a $\sim$50\% latency improvement compared to the next best FSL SoTA baseline on VQA and OKVQA datasets.


\appendix



\bibliographystyle{unsrt}
\bibliography{main}

\end{document}